%% file: wacv-2026-author-kit-template/main.tex
\documentclass[10pt,twocolumn,letterpaper]{article}

\usepackage[pagenumbers]{wacv} %
\usepackage{siunitx}
\usepackage{pifont}
\newcommand{\cmark}{\ding{51}} %
\newcommand{\xmark}{\ding{55}} %

\usepackage{xspace}
\newcommand{\RLow}{\textsf{R128}\xspace}
\newcommand{\RMid}{\textsf{R256}\xspace}
\newcommand{\RHigh}{\textsf{R512}\xspace}

\newcommand{\Dimlow}{$\num{128}\times\num{32}\times\num{32}$\xspace}
\newcommand{\Dimmid}{$\num{256}\times\num{64}\times\num{64}$\xspace}
\newcommand{\Dimhigh}{$\num{512}\times\num{128}\times\num{128}$\xspace}

\newcommand{\PMRT}{PMRT\xspace}
\newcommand{\PMRTExpanded}{Progressive Multi-Resolution Training\xspace}

\newcommand{\drivaermlcite}{2}
\newcommand{\drivaernetppcite}{11}
\newcommand{\korinacite}{32}
\newcommand{\streamingcite}{40}


\definecolor{wacvblue}{rgb}{0.21,0.49,0.74}
\usepackage[pagebackref,breaklinks,colorlinks,allcolors=wacvblue]{hyperref}
\usepackage{float}

\def\wacvPaperID{2485} %
\def\confName{WACV}
\def\confYear{2026}

\title{PMRT: A Training Recipe for Fast, 3D High-Resolution Aerodynamic Prediction}

\author{Sam Jacob Jacob \\
Volkswagen AG\\
Wolfsburg, Germany\\
{\tt\small sam.jacob.jacob@volkswagen.de}
\and
Markus Mrosek \\
Volkswagen AG\\
Wolfsburg, Germany\\
{\tt\small markus.mrosek@volkswagen.de}
\and
Carsten Othmer \\
Volkswagen AG\\
Wolfsburg, Germany\\
{\tt\small carsten.othmer@volkswagen.de}
\and
Harald Köstler\\
Friedrich-Alexander-Universität Erlangen-Nürnberg\\
Erlangen, Germany\\
{\tt\small harald.koestler@fau.de}
}

\begin{document}
\maketitle

\input{Figures/src/overview_plot}

\begin{abstract}
The aerodynamic optimization of cars requires close collaboration between aerodynamicists and stylists, while slow, expensive simulations remain a bottleneck. Surrogate models have been shown to accurately predict aerodynamics within the design space for which they were trained. However, many of these models struggle to scale to higher resolutions because of the 3D nature of the problem and data scarcity. We propose \PMRTExpanded (\PMRT), a probabilistic multi-resolution training schedule that enables training a U-Net to predict the drag coefficient ($c_d$) and high-resolution velocity fields (\Dimhigh) in \SI{24}{\hour} on a single NVIDIA H100 GPU, 7× cheaper than the high-resolution-only baseline, with similar accuracy. \PMRT samples batches from three resolutions based on probabilities that change during training, starting with an emphasis on lower resolutions and gradually shifting toward higher resolutions. Since this is a training methodology, it can be adapted to other high-resolution-focused backbones. We also show that a single model can be trained across five datasets from different solvers, including a real-world dataset, by conditioning on the simulation parameters. In the DrivAerML dataset, our models achieve a $c_d$ $R^2$ of 0.975, matching literature baselines at a fraction of the training cost.
\end{abstract}

\section{Introduction}
\label{sec:intro}

Improving the aerodynamics of a car is essential, as it improves fuel economy throughout the vehicle’s lifetime and is even more paramount for electric vehicles, where it increases range for the same battery capacity. Typically, the aerodynamic optimization process consists of iterations between aerodynamicists and stylists as both the aerodynamics and design of the car are essential. However, a major bottleneck are the time-consuming CFD simulations, which take on average \SI{20}{\hour} and up to several days on~\num{1000} CPU cores. This bottleneck leads to fewer iterations between aerodynamicists and stylists or reduced exploration of the design space. Aerodynamicists could use surrogate models that enable real-time aerodynamic predictions to explore the design space before committing to testing a few promising variations using expensive simulations.

\subsection{Related work}
Existing studies have successfully applied surrogate models for various general fluid dynamics applications \cite{Survery.General.Lino.2023, Survey.General.Brunton.2020} and automotive aerodynamic applications \cite{Jacob_SAE_DrivAer, BenchmarkNvidia, AB-UPT, DoMINO, XMeshGraphNetsNvidia}. There are several approaches to creating surrogate models, and we highlight a few subsequently. Convolutional neural network (CNN)-based methods typically operate on intermediate volumetric encodings (instead of meshes) like binary masks \cite{BinaryMask.Thuerey.2020, BinaryMask.ScalableCNN.Rana.2024}, signed distance fields (SDFs) \cite{Jacob_SAE_DrivAer, Jacob_benchmarking_CNN_GNN,SDF.CNN.Chen.2021,SDF.Bhatnagar.2019,SDF.Guo.2016}, or more recently alternative encodings like triplanes \cite{TripNet.Chen.3192025} and factorized implicit grids \cite{FIGConvNvidia}. Earlier CNN-based approaches typically did not predict surface fields; recent approaches \cite{FIGConvNvidia, TripNet.Chen.3192025} address this. Graph-based \cite{XMeshGraphNetsNvidia, GNN.Pfaff., GNN.SanchezGonzalez.2020} and neural-operator-based \cite{DoMINO, AB-UPT, GINO.Li.2023} models work natively on meshes or point clouds, avoiding volumetric intermediates. One of the persistent challenges with surrogate models has been training at higher spatial resolutions. Several methods use low-resolution voxels or significantly downsampled meshes. Higher resolutions can improve geometry representation and enable the prediction of finer flow details. In 3D, memory and compute scale cubically, and many otherwise strong methods struggle to scale or become computationally expensive. Compounding this, high-fidelity aerodynamics simulations are scarce due to the high computational cost. Recently, several works have addressed this problem in automotive aerodynamics, including: scalable GNN \cite{XMeshGraphNetsNvidia} (partitioned multi-scale graphs with halo exchange); scalable neural operators \cite{AB-UPT, DoMINO} (anchored branched universal physics transformers; decomposable neural operator); alternative representations like triplane \cite{TripNet.Chen.3192025}; factorized implicit grids \cite{FIGConvNvidia}; and super-resolution models \cite{3D_super_resolution_Thanh} that predict high-resolution fields from coarse fields.

In parallel, there have been various high-resolution vision strategies: these include fine-tuning/curriculum training at higher resolutions \cite{FineTuneHighRes.SwinV2,Finetuning.highres.Hugo,FinetuningHighRes.Howard.July2020,ProgressiveGrowingGan.Karras.10272017,FinetuningHighRes.Touvron.6152019,ProgressiveResolution.EfficientNetV2.Tan.}; mixed-resolution/patch-size and aspect-agnostic training \cite{FlexVit.Beyer.12152022,NaViT.Dehghani.7122023,ResFormer.Tian.1212022}; zoom-in pyramids \cite{Dragonfly.Thapa.632024,SNIPER.Singh.5232018}; backbone improvements \cite{FineTuneHighRes.SwinV2,HRNet.HighResBackbone}; sparsity \cite{Sparse.Williams.2024,Sparse.Wang.2017} and token reduction \cite{Droptoken.Akbari.4222021,Droptoken.Yin.12142021}. However, porting these ideas to 3D CFD surrogates is not always trivial; for example: in 3D memory scales cubically with voxel size; required resolution jumps are significant (e.g., from \Dimlow to \Dimhigh); sharp switches in resolution could destabilize training; scale-based augmentations that help vision tasks can occlude geometry, and even change the actual physical targets; vision transformers, even with these improvements, can be too expensive in cases where time-to-first prediction is critical. Nevertheless, these ideas inspire our training recipe.

\subsection{Gap and Overview}

Even with the above advances, we still lack a training-only recipe inspired by mixing-resolutions or curriculum-style training that scales proven voxel-based CNN approaches to high-resolution field prediction without significant architectural changes. Rather than abandoning CNNs for high-resolution-focused backbones, we adapt the training and propose \PMRTExpanded (\PMRT - see \cref{fig:prt_overview_plot} for an overview): a smooth, probabilistic schedule that mixes resolutions during training, starting with an emphasis on lower resolutions and gradually transitioning to higher resolutions within a single run, avoiding hard resolution switches and enabling accurate high-resolution volumetric predictions with low compute budgets.

The availability of aerodynamic simulations is limited by high compute cost and significant human effort; for example, the estimated compute cost for the DrivAerML dataset \cite{DrivAerML} (500 simulations) is $\approx\$1M$ \cite{AB-UPT}. With the currently available data, it is not yet feasible to train a generalizable model that accurately predicts the aerodynamics of any car geometry, but surrogate models have been shown to generalize well within the design space on which they were trained; this already helps aerodynamicists. As a result, surrogate models must be retrained for new projects or for baselines that differ substantially from existing geometries.

Even with high-fidelity simulations, it is essential to validate them several times against wind-tunnel experiments during vehicle development. Similarly, the primary application of surrogate models we evaluate in this study is to explore the design space and then validate promising solutions using simulations. Therefore, when the training time is too high, it may no longer make sense to use surrogate models, as a short time-to-first prediction is crucial on new datasets. With lower training time, aerodynamicists can utilize additional data or a new dataset to train models overnight or within a day. For these reasons, we impose a one-day training budget for our models. Finally, we train two model families: $c_d$-only and $c_d$ + velocity fields, since velocity fields are often unavailable (not archived due to storage costs: raw DrivAerML dataset $\approx\SI{31}{\tera\byte}$) or unnecessary for a given study, and they increase training time. Our main contributions include:
\begin{itemize}
    \item \PMRTExpanded (\PMRT): a probabilistic multi-resolution schedule that enables accurate high-resolution velocity field prediction with low training costs (single NVIDIA H100 GPU: $\approx$ \SI{24}{\hour} for all datasets, $7\times$ faster than training only on high-resolution; $\approx$\SI{4}{\hour} for DrivAerML). It also improved $c_d$ prediction accuracy for $c_d$-only models using high-resolution SDF inputs.
    \item Competitive DrivAerML accuracy with a U-Net, at substantially lower wall-clock time compared to methods reported in the literature.
    \item A single U-Net conditioned on simulation parameters trained jointly on five datasets ($\approx1900$ training simulations) created using different solvers.
\end{itemize}

\section{Datasets}

\input{Figures/src/drag_distribution}

\begin{table*}
  \centering
  \begin{tabular*}{\textwidth}{@{\extracolsep{\fill}}llcccccl@{}}
    \toprule
    Dataset & Group &
    \begin{tabular}[c]{@{}c@{}}Parametric\\ (\cmark/\xmark)\end{tabular} &
    \begin{tabular}[c]{@{}c@{}}No. morphing\\ parameters\end{tabular} &
    \begin{tabular}[c]{@{}c@{}}No. training\\ samples\end{tabular} &
    \begin{tabular}[c]{@{}c@{}}No. test\\ samples\end{tabular} &
    \begin{tabular}[c]{@{}c@{}}No. baseline\\ geometries\end{tabular} &
    Solver \\
    \midrule
    DrivAerML    & DrivAerML  & \cmark & 16 & 411 & 73  & 1  & OpenFOAM \\
    DrivAer      & Parametric & \cmark & 15 & 850 & 150 & 1  & ultraFluidX \\
    Jetta        & Parametric & \cmark & 6  & 85  & 15  & 1  & ultraFluidX \\
    Parametric I & Parametric & \cmark & 7  & 262 & 46  & 1  & OpenFOAM \\
    Real-world   & Real-world & \xmark & -- & 319 & 55  & 39 & OpenFOAM \\
    \bottomrule
  \end{tabular*}
  \caption{Summary of the datasets used. We indicate if the dataset is parametric with \cmark/\xmark, along with the number of morphing parameters, training / test samples, baseline geometries, and solver. We group the datasets into DrivAerML (public benchmark), Parametric (parametric changes, single baseline), and Real-world (free-form, multiple baselines), to enable easier comparison and balanced evaluation.}
  \label{tab:datasets}
\end{table*}

We use five datasets, which can be classified into two types: parametric and real-world (or non-parametric). Parametric datasets are generated by morphing a baseline geometry for a set of chosen geometric parameters within predefined ranges; real-world datasets are generated from free-form deformations applied across multiple baselines without explicit parameterization. We refer to the latter as 'real-world' because they are similar to the datasets curated from simulations generated during vehicle development without being explicitly created for training models.

\Cref{tab:datasets} gives an overview of the datasets. DrivAer \cite{DrivAer.Markus} and DrivAerML \cite{DrivAerML} datasets are parametric datasets based on the publicly available DrivAer geometry introduced by Heft \etal \cite{DrivAer.Heft.2012}. DrivAerML \cite{DrivAerML} is a publicly available parametric dataset of 484 simulations with 16 morphing parameters, simulated using OpenFOAM. DrivAer \cite{DrivAer.Markus} is a parametric dataset of 1,000 simulations with 15 morphing parameters. Jetta \cite{ROM.Jetta.Mrosek.2019} is a parametric dataset of 100 simulations with 6 morphing parameters. The DrivAer and Jetta datasets were simulated using Altair's ultraFluidX, a lattice Boltzmann solver, following the simulation setup described by Mrosek \etal \cite{DrivAer.Markus}. Parametric I and real-world datasets are internal datasets created by an automotive manufacturer; the simulations were performed using OpenFOAM following the setup by Islam \etal \cite{Islam.2009.OpenFOAMSetup}. Parametric I is a parametric dataset comprising 308 simulations with seven morphing parameters. The real-world dataset is a non-parametric dataset comprising 374 simulations with 39 baseline geometries (Sedan or CUV). For all datasets, we use time-averaged $c_d$ and velocity fields. Two types of time-averaging are employed in the datasets: (1) classic averaging, where simulations are run for a total of \SI{4}{\second} of physical time, and the final \SI{2}{\second} are averaged; and (2) a more recent dynamic averaging (using the tool Meancalc), where the total simulated physical time (with a maximum cap), start and end of the averaging window are dynamically adjusted to achieve a target statistical accuracy. DrivAer, Jetta, and Parametric I use classic averaging. DrivAerML \cite{DrivAerML} uses dynamic averaging to reach $\pm1.5$ drag counts accuracy. Approximately one-third of real-world simulations use classic averaging, while the rest use dynamic averaging to reach $\pm0.5$ drag counts accuracy.

\Cref{fig:drag distribution} shows the distribution of $c_d$ across the training and test partitions. We generate the splits using stratified splitting based on $c_d$ and morphing parameters for the parametric datasets, and on $c_d$ and the baseline group (group of vehicles derived from the same baseline geometry) for the real-world dataset. For the validation split, we use \SI{15}{\percent} of the samples from the training partition. Refer to the supplementary material for further evaluation of the datasets.

\section{Methodology}

\subsection{Preprocessing}

We predict the velocity field on a Cartesian grid in a region of interest (ROI) around the car. The ROI is chosen (DrivAerML ROI in supplementary) based on the region where most of the interesting features in the velocity field occur, and we interpolate the simulated velocity fields onto the Cartesian grid in the ROI. Cars are typically represented as meshes: an unstructured form of geometry representation that cannot be directly passed to CNNs. We generate SDFs for these geometries in a Cartesian grid in the same ROI where the velocity field is predicted. Each SDF value is the shortest distance from the voxel (center) to the surface of the geometry, with the sign being positive for the voxels outside the geometry and negative otherwise. We generate the SDFs using OpenVDB \cite{openVDB}. During training, we apply the same data augmentation used by Jacob \etal \cite{Jacob_benchmarking_CNN_GNN}.

Different solvers and setup choices introduce variability; to manage this, we pass simulation parameters as an additional input: these include the simulation tool used and the setup parameters (for example, simulated physical time). We normalize (to have zero mean and unit variance) the simulation parameters and then reduce the dimensionality using principal component analysis before passing them to the model. The SDF and velocity fields are also normalized to have zero mean and unit variance. We use three voxel (grid) resolutions: \Dimlow (\RLow), \Dimmid (\RMid), and \Dimhigh (\RHigh). We train the models to predict at \RLow or \RHigh, and use \RMid only during pre-training.

\subsection{\PMRTExpanded (PMRT)}

Training models using high-resolution SDFs for aerodynamic applications is computationally expensive and challenging due to data scarcity. To address this, we propose \PMRT, a probabilistic multi-resolution training schedule that balances compute and accuracy. During training, we dynamically sample batches from multiple resolutions based on probabilities that vary throughout the training process. The schedule emphasizes lower resolutions at the start and gradually transitions to higher resolutions. In contrast to sharp resolution switches, the \PMRT schedule helps to stabilize training, and since scale-based augmentations are not applicable, a per-resolution probability floor keeps all resolutions sampled for most of training, encouraging multiscale feature learning.

We use a three-phase schedule (warm-up, pre-training, and fine-tuning) to train on the \RLow, \RMid, and \RHigh SDFs. \Cref{fig:prt_overview_plot} illustrates an example \PMRT schedule and shows how it affects training costs. The warm-up phase begins with equal probabilities across all resolution levels and linearly interpolates to the starting probabilities of the pre-training phase. During the pre-training phase, we gradually transition the sampling probabilities from emphasizing lower resolutions early in training to higher resolutions later in training. The pre-training schedule is generated by discretizing a Gaussian distribution over the resolution indices, where the mean moves from low toward higher resolutions while the standard deviation shrinks. To prevent neglecting any resolution, we enforce the per-resolution probability floor by clipping the raw sampling probabilities at a minimum value. At the end of the pre-training phase for five epochs, we linearly interpolate the probabilities from the mixed distribution probabilities to the fine-tuning probabilities (only the high resolution). During warm-up and pre-training, we multiply the batch size by a factor of 4 for all the lower resolutions. Finally, we have the fine-tuning phase, where the model continues training for additional epochs using only the highest resolution. The warm-up phase slightly improves the stability of the training and convergence. Pre-training accounts for most of the convergence; fine-tuning specializes the model to \RHigh. See the supplementary material for the implementation details and ablations of \PMRT design choices.

\subsection{Architecture}

We use a U-Net \cite{UNet.base} closely following Jacob \etal \cite{Jacob_SAE_DrivAer}. The encoder extracts features from the SDF, which are concatenated with the processed simulation parameter input to form the bottleneck vector. From this bottleneck, three identical decoders predict the three velocity components, and a prediction head predicts $c_d$. The architecture is modular and can be extended for additional volumetric fields and coefficients. For $c_d$-only models, we omit the velocity decoders.

We pass the SDF input through an input convolution block followed by six encoder blocks (concurrent spatial \& channel squeeze \& excitation layer \cite{SqueezeAndExcitation.Hu.2018, SqueezeAndExcitation.Roy.2018} \(\rightarrow\) PReLU activation \cite{PRELU.He.2015} \(\rightarrow\) convolution layer \(\rightarrow\) max-pooling (with stride 2) \(\rightarrow\) group normalization \cite{GroupNormalization.Wu.2018} with residual skip connection) and finally a global pooling layer. The simulation parameter input is passed through an MLP block (linear layer \(\rightarrow\) activation \(\rightarrow\) dropout \(\rightarrow\) normalization) to expand the simulation parameters to have the same shape as the encoder's output, followed by a multi-head self-attention layer. The resulting vector and the encoder output are then concatenated to get the bottleneck vector. 

The bottleneck vector is passed through three U-Net decoders, one for each component of velocity, which mirrors the encoder and has six decoder blocks followed by an output convolution layer. Each decoder block has a transpose convolution layer instead of the convolution and pooling layer. In addition to the standard skip connections present in the U-Net architecture, we introduce a direct skip from the input SDF to the output convolution: we compute a binary mask from the SDF, concatenate the SDF and mask, pass it through two convolution blocks (convolution \(\rightarrow\) normalization \(\rightarrow\) activation) to match channels, and then concatenate this feature map before the output convolution layer.

We pass the bottleneck vector through a coefficient prediction head (transformer layer \(\rightarrow\) MLP block \(\rightarrow\) linear layer) to predict $c_d$. The transformer layer allows the model to learn the relationships between the simulation parameters and the features extracted from the encoder. 

To support multiple resolutions, we apply global pooling at the end of the encoder to produce a fixed-size output. In the decoder, we use nearest-neighbor interpolation to align tensor sizes when they do not match in the first decoder block. For the $c_d$-only models, we use a batch size of 64, and for $c_d$ + velocity field models, we use a batch size of 16. For training, we use a cyclic learning rate \cite{CLR.Smith.2017} with the NAdam optimizer and apply stochastic depth regularization \cite{StochasticDepth.Huang.2016}. We present ablations on our architectural choices in the supplementary material.

\subsection{Loss function}

We use a Smooth L1 loss for both $c_d$ and velocity. We build a volumetric loss weighting from two complementary strategies: distance-based (Trinh~\etal \cite{3D_super_resolution_Thanh}) and variance-based (Jacob~\etal~\cite{Jacob_SAE_DrivAer}) weighting. Distance-based weighting assigns higher weights to voxels near the vehicle surface, which decay with distance. Variance-based weighting is precomputed for a dataset based on the cell-wise velocity field variance across the training set; that is, cells with higher velocity field variance receive larger weights, typically those near the surface and in the wake. 

\begin{equation}
w =
\mathcal{N}\Big[
\mathcal{N}\big( c + \tfrac{\alpha \cdot \mathrm{mask}}{\alpha + \mathrm{uSDF}} \big)
+
\mathcal{N}\big( \lVert \nabla U \rVert_{2}^{1/4} \big)
\Big]
\label{eq:total_velocity_weight}
\end{equation}

\Cref{eq:total_velocity_weight} represents the combined weighting function. $(\tfrac{\alpha \cdot \mathrm{mask}}{\alpha + \mathrm{uSDF}})$ is the distance-based weighting from Trinh~\etal \cite{3D_super_resolution_Thanh}. The $\mathrm{mask}$ is 1 for regions with fluid flow and 0 otherwise, and $\mathrm{uSDF}$ is the unsigned distance to surface. We add a constant $c$ to the distance-based weighting so that the cells inside the vehicle geometry have a non-zero weight. This ensures the model is penalized for predicting non-zero velocities inside the geometry. $\alpha$ and $c$ are constants corresponding to 5.5 (from Trinh et al. \cite{3D_super_resolution_Thanh}) and 0.75. $\mathcal{N}[\cdot]$ denotes normalization so that the component has a unit mean for the sample. For $c = 0.75$, after normalizations, interior voxels typically have a weight $\approx$0.25 in representative samples. To avoid precomputing variances, we use the gradient of the velocity magnitude ($\lVert \nabla U \rVert$ is the L2 norm of the gradient of the velocity magnitude). The gradient of velocity magnitude is higher close to the vehicle's surface and in the wake, but it can spike near the vehicle's surface, in some regions of the wake, and contains some noise. To obtain a smoother weight, we raise the gradient magnitude to a power of 0.25; the exponent was empirically chosen to preserve the goal of the weighting. To ensure the two terms contribute equally, we normalize each term, sum them, and then renormalize.

\begin{table*}[h]
  \centering
  \begin{tabular*}{0.99\textwidth}{@{\extracolsep{\fill}}lccccccc@{}}
    \toprule
    \begin{tabular}[c]{@{}c@{}}Variant\\ (resolution)\end{tabular} &
    \begin{tabular}[c]{@{}c@{}}Velocity\\ (\cmark/\xmark)\end{tabular} &
    \begin{tabular}[c]{@{}c@{}}\PMRT\\ (\cmark/\xmark)\end{tabular} &
    \begin{tabular}[c]{@{}c@{}}Pre-training\\ epochs\end{tabular} &
    \begin{tabular}[c]{@{}c@{}}Fine-tuning\\ epochs\end{tabular} &
    \begin{tabular}[c]{@{}c@{}}Group:\\ $c_d$ MAE $\downarrow$\end{tabular} &
    \begin{tabular}[c]{@{}c@{}}Group: relative\\ L2 error $\downarrow$ [\si{\percent}] \end{tabular} &
    \begin{tabular}[c]{@{}c@{}}Wall-clock\\ time [\si{\hour}] \end{tabular} \\
    \midrule
    \RLow \ddag        & \xmark & \xmark & --  & --  & 3.4  & --  & 3   \\
    \RHigh          & \xmark & \xmark & --  & --  & \bfseries  3.0 & --  & 24  \\
    \RHigh *        & \xmark & \xmark & --  & --  & 3.0 & --  & 66 \\
    \cmidrule(lr){1-8}
    \RHigh          & \xmark & \cmark & 200 & 50  & 2.8  & --  & 5   \\
    \RHigh \ddag        & \xmark & \cmark & 400 & 50  & \bfseries 2.5 & --  & 8  \\
    \RHigh          & \xmark & \cmark & 200 & 400 & 2.5 & --  & 16  \\
    \RHigh          & \xmark & \cmark & 400 & 400 & 2.5 & --  & 18  \\
    \cmidrule(lr){1-8}
    \RLow \ddag        & \cmark & \xmark & --  & --  & \bfseries 2.4 & 2.7 & 9   \\
    \RHigh *        & \cmark & \xmark & --  & --  & 2.7  & 2.6  & 72   \\
    \RHigh *        & \cmark & \xmark & --  & --  & \bfseries  2.4  & \bfseries 2.4  & 168  \\
    \cmidrule(lr){1-8}
    \RHigh          & \cmark & \cmark & 50  & 50  & 2.9 & 2.7 & 18  \\
    \RHigh \ddag        & \cmark & \cmark & 50  & 75  & \bfseries 2.5 & \bfseries 2.4 & 24 \\
    \bottomrule
  \end{tabular*}
  \caption{Test metrics for models trained on all datasets. Each row is a variant with resolution \RLow or \RHigh, trained with/without velocity and with/without \PMRT. We report pre-training and fine-tuning epochs, group metrics, and training time on a single NVIDIA H100 GPU. The best models selected for the downstream tables are marked \ddag; rows marked * are \RHigh models without \PMRT that exceed the \SI{24}{\hour} budget. The \PMRT \RHigh models compared to baseline have similar or better accuracy at significantly lower training costs.}
  \label{tab:main_results}
\end{table*}

\input{Figures/src/velocity_field_prediction_comparision}

\subsection{Benchmark \& Evaluation}

Generating simulations exclusively for training models is expensive, and to sustainably train in the future, we hope most training data comes from archived simulations from past vehicle development projects. These simulations likely use different simulation parameters and solvers; to show that it is possible to train a single model on diverse datasets and to encourage learning shared features that transfer across datasets, we train a single model jointly on all datasets. We report group-averaged metrics on five datasets to enable easier comparison between models. We partition the datasets into three groups (\cref{tab:datasets} - DrivAerML (public benchmark), Parametric (parametric changes, single baseline), and Real-world (free-form, multiple baselines)) and compute metrics within each group. We then average the group metrics so each group contributes equally. The only multi-dataset group is 'Parametric'; since the group has a disproportionately large number of DrivAer samples, we average across datasets within this group so each dataset contributes equally. For the $c_d$ prediction, we report the Mean Absolute Error (MAE) and the Maximum Absolute Error (MaxAE; supplementary only) in drag counts\footnote{One drag count = 0.001 drag.}.  

Since DrivAerML is the only publicly available dataset used in our study, we retrain the best models using only the DrivAerML dataset based on the splits used in the benchmark study by Tangsali \etal \cite{BenchmarkNvidia}; the same split is used by all the models compared. The split has a total of 48 test samples, and \SI{20}{\percent} of the test samples are out-of-distribution samples. We compare our best models with AB-UPT~\cite{AB-UPT}, X-MeshGraphNet~\cite{XMeshGraphNetsNvidia,BenchmarkNvidia}, FIGConvNet~\cite{FIGConvNvidia,BenchmarkNvidia} and DoMINO~\cite{DoMINO,BenchmarkNvidia} on the DrivAerML~\cite{DrivAerML} dataset; hereafter, we refer to these models as the literature baselines. Since some literature baselines report only the drag force $R^2$ score, we report the drag force $R^2$ score and $c_d$ $R^2$ when comparing with the literature baselines. For velocity field prediction, we report the relative L2 error:

\begin{equation}
    \text{Rel. L2 Error} = \left( \frac{1}{n_{\text{test}}} \sum_{i=1}^{n_{\text{test}}}
    \frac{\left\| \mathbf{u}^i - \hat{\mathbf{u}}^i \right\|_2}{\left\| \mathbf{u}^i \right\|_2} \right) \times 100 \, [\%]
    \label{eq:l2_percent}
\end{equation}

\noindent where \( n_{\text{test}} \) is the number of test samples; \( \mathbf{u}^i \) and \( \hat{\mathbf{u}}^i \) are the true and predicted velocity field vectors, respectively.

\section{Results}

\begin{table*}[t]
  \centering
  \begin{tabular*}{\textwidth}{@{\extracolsep{\fill}}lccccc@{}}
    \toprule
    \begin{tabular}[c]{@{}c@{}}Method\end{tabular} &
    \begin{tabular}[c]{@{}c@{}}Velocity\\ (\cmark/\xmark)\end{tabular} &
    \begin{tabular}[c]{@{}c@{}}$c_d$: $R^2$\\ score $\uparrow$\end{tabular} &
    \begin{tabular}[c]{@{}c@{}}Drag force:\\ $R^2$ score $\uparrow$\end{tabular} &
    \begin{tabular}[c]{@{}c@{}}Velocity: relative L2\\ error [\%] $\downarrow$ $^{\ast}$\end{tabular} &
    \begin{tabular}[c]{@{}c@{}}NVIDIA H100 \\ GPU-hours\\ \end{tabular} \\
    \midrule
    AB-UPT \dag                & \cmark & 0.976               & 0.992               & 5.99           & 7  \\
    DoMINO                 & \cmark & --                  & 0.984               & --             & 184 \ddag \\
    FIGConvNet             & \xmark & --                  & 0.975               & --             & -- \\
    X-MeshGraphNet         & \cmark & --                  & 0.921               & --             & -- \\
    \cmidrule(lr){1-6}
    Ours: \RLow   & \xmark & $0.952 \pm 0.014$   & $0.986 \pm 0.004$   & --             & 1  \\
    Ours: \RLow   & \cmark & $0.970 \pm 0.003$   & $0.991 \pm 0.001$   & $3.7 \pm 0.1$  & 1  \\
    Ours: \PMRT \RHigh   & \xmark & $0.975 \pm 0.002$   & $0.993 \pm 0.001$   & --             & 1  \\
    Ours: \PMRT \RHigh   & \cmark & $0.974 \pm 0.002$   & $0.992 \pm 0.001$   & $2.4 \pm 0.1$  & 4  \\
    \bottomrule
  \end{tabular*}
  \caption{DrivAerML benchmark split evaluation \cite{BenchmarkNvidia}: comparison with literature baselines. To keep the evaluation fair, our models are trained only on DrivAerML. We report $c_d$ $R^2$, drag force $R^2$, velocity relative L2 error [\%], and compute cost in GPU-hours (1$\times$ NVIDIA H100). Notes: “Ours” shows mean ± std over five runs; \dag~AB-UPT shows the median over five runs for which the author provided the metrics; \ddag~GPU-hours for DoMINO were provided by the authors; $^{\ast}$ the metric is not comparable across methods due to different prediction regions; all other values are reproduced from the benchmark study \cite{BenchmarkNvidia} where some of them are not reported.}
  \label{tab:results models DrivAerML DoMINO split}
\end{table*}

\input{Figures/src/drag_correlation_plot}

All models in this study are trained on a single NVIDIA H100 GPU. We start with $c_d$-only models, followed by $c_d$+velocity field models trained on all the datasets. Finally, we compare our models trained only on DrivAerML with literature baselines. For \PMRT, we always use a probability floor of 0.1 with 10 warm-up epochs. Extended subgroup test metrics are reported in the supplementary material.

\subsection{Model predicting only drag coefficient}

From the results in \cref{tab:main_results}, we can see that the baseline \RLow model achieves a group $c_d$ MAE of 3.4 drag counts with a wall-clock time of \SI{3}{\hour}. Training the model at \RHigh without \PMRT reduces the error by 0.4 drag counts, with an 8 times increased training time of \SI{24}{\hour}. Extending the training does not improve accuracy.

With \PMRT, the cheapest \RHigh model attains a group $c_d$ MAE of 2.8 drag counts in \SI{5}{\hour}, outperforming all baselines with only a \SI{2}{\hour} increase in training cost. We ablate the pre-training epochs (200, 400) and fine-tuning epochs (50, 400). The model with the best balance of accuracy and compute cost is the \PMRT \RHigh model with 400 pre-training and 50 fine-tuning epochs, which lowers MAE by \SI{26}{\percent} vs \RLow and \SI{17}{\percent} vs \RHigh without \PMRT while training three times faster (\SI{8}{\hour}). The \PMRT \RHigh models with 400 fine-tuning epochs do not improve accuracy beyond 2.5 drag counts, but incur substantially higher training costs (\SIrange{16}{18}{\hour}).

Pre-training and fine-tuning epochs are hyperparameters that should be chosen based on experimentation. In our experiments, longer pre-training with a short fine-tuning achieved the same accuracy as longer fine-tuning, with lower wall-clock time. For a fixed budget, shifting epochs to pre-training is more compute-efficient.

\subsection{Model predicting drag coefficient and velocity field}

The baseline \RLow $c_d$ + velocity model (correlation plot in \cref{fig:correlation plot}, left \& middle) attains a group $c_d$ MAE of 2.4 drag counts and a group relative L2 error of \SI{2.7}{\percent} with a wall-clock time of \SI{9}{\hour}. We attribute the improvement compared to \RLow $c_d$-only model to additional supervision from jointly predicting $c_d$ and velocity fields, which increases the information available during training. Without \PMRT, \RHigh baselines do not reach a good solution within \SI{24}{\hour}. When trained for 3 days, the model achieves a group $c_d$ MAE of 2.7 drag counts; extending the training time to 7 days finally matches the \RLow $c_d$ MAE and improves the group relative L2 error by \SI{11}{\percent} at $\approx19\times$ the training time of \RLow. The best \PMRT \RHigh model achieves a group $c_d$ MAE of 2.5 drag counts and a group relative L2 error of \SI{2.4}{\percent} in \SI{24}{\hour}, delivering similar accuracy to the \RHigh without \PMRT model while being $7\times$ cheaper to train.

For $c_d$ prediction, the \RLow $c_d$ + velocity baseline model outperforms the \RLow $c_d$-only baseline. While the \PMRT \RHigh $c_d$-only model and the \RLow $c_d$ + velocity baseline model have similar $c_d$ accuracy with similar wall-clock time (8 and \SI{9}{\hour}); the choice depends on whether fields are needed. If fields are unnecessary or unavailable, training the \PMRT \RHigh $c_d$-only is a viable option. 

By visually comparing \RLow and \PMRT \RHigh velocity field predictions in \cref{fig:velocity field comparision}, we observe finer spatial detail captured by the \RHigh model. For example, even in the \RLow ground truth, some details are lost in the wake, which are present in the \RHigh ground truth, and many of these details are also captured by the \RHigh model’s prediction. This presents a clear trade-off: we recommend \PMRT \RHigh model, as it provides higher-detail velocity fields with lower relative L2 error. However, with limited compute, the \RLow model remains a practical choice for accurate $c_d$ with coarse velocity field prediction. For both models, errors are concentrated in similar regions: near the vehicle surface and in the wake. 

We observe that the $c_d$ MAE plateaus at 2.4 drag counts, achieved by the \RLow $c_d$ + velocity model; we hypothesize a few reasons for the lack of $c_d$ prediction improvement in the \RHigh models: (1) a fundamental limit due to simulation noise that can distort physical trends; for example, the $c_d$ and velocity fields are time-averaged; the DrivAerML simulations were set up for a statistical accuracy of $\pm$1.5 drag counts; this is already close to what our models and other SOTA models achieve; (2) architectural limitations: we use a U-Net that has been shown to work for low resolution aerodynamic applications, further accuracy improvements might need architectural improvements; (3) despite its coarseness, \RLow may still encode sufficient geometric information for accurate $c_d$ prediction.

\subsection{Benchmarking on DrivAerML dataset}

For a fair comparison with existing models, we train our models only on the DrivAerML dataset using the split used in the benchmark study by Tangsali \etal \cite{BenchmarkNvidia}. The results can be seen in \cref{tab:results models DrivAerML DoMINO split}, some metrics are not available because the literature baselines did not report them. We train our model five times with different seeds and report the mean and standard deviation of the metrics. Among the literature baselines, AB-UPT has the best accuracy, followed by DoMINO. Except for the \RLow $c_d$-only model, our $c_d$ and drag force $R^2$ scores are similar to AB-UPT. \Cref{fig:correlation plot} (right-most) shows our model captures the trend well. For the velocity field, our model has a lower relative L2 error compared to AB-UPT. However, the relative L2 error metrics are not comparable because the methods predict the fields in different regions, and our method predicts on a Cartesian grid; the literature baselines predict on unstructured grids.

Training on DrivAerML takes $\approx$\SI{4}{\hour} for the \PMRT \RHigh $c_d$ + velocity field model and $\approx$\SI{1}{\hour} for others, compared to \SI{7}{\hour} for AB-UPT. While conventional CNN-based models, like ours, have limitations (e.g., the inability to directly predict surface fields and dependence on resolution), methods such as AB-UPT and DoMINO address these. Despite these limitations, our models strike a balance due to their lower compute cost, making them suitable for scenarios requiring short time-to-first prediction on new datasets or with limited compute or when fields are unavailable. Moreover, we have shown that our models can scale to multi-dataset scenarios using different solvers while keeping the compute cost reasonable.

\section{Conclusion}
In this study, we address the problem of high computational cost in training CNN-based surrogates at high resolutions for automotive aerodynamics. We present \PMRT, a probabilistic multi-resolution training schedule that enables high-resolution training at substantially lower cost without significant architectural changes. With \PMRT, we can train a U-Net on five datasets ($\approx$ 1900 samples) for high-resolution field prediction in \SI{24}{\hour} on a single NVIDIA H100 GPU, up to $7\times$ cheaper than our high resolution without \PMRT model, while matching or improving accuracy. We demonstrate that conditioning with simulation parameters enables us to train a single model on datasets from different solvers, including a real-world dataset. Our models also achieve similar accuracy compared to literature baselines at a fraction of the training cost on the DrivAerML dataset. Since \PMRT is a training methodology, we leave its evaluation on other high-resolution-focused backbones to future work.

\nocite{Kornia, mosaicml2022streaming, DrivAerNet++}

\section*{Acknowledgments}
The authors acknowledge the assistance of colleagues who collected the dataset and helped us throughout the study. 

\section*{Disclaimer}
The results, opinions and conclusions expressed in this publication are not necessarily those of Volkswagen Aktiengesellschaft.


\input{main.bbl}
\clearpage  
\onecolumn  
\appendix

\section*{Supplementary}

\section{Comparison of datasets}

\input{Figures/src/dataset_comparision}

Currently, the only publicly available high-fidelity aerodynamics dataset is the DrivAerML [\drivaermlcite] dataset. Although DrivAerNet++ [\drivaernetppcite] is a large aerodynamics dataset, its simulations are based on Reynolds-Averaged Navier-Stokes (RANS), which carry higher errors compared to the standard high-fidelity simulations in the industry. In this study, we do not evaluate DrivAerNet++ due to its simulation method and licensing restrictions. Both datasets are parametric and have greatly helped improve aerodynamic surrogates. However, we currently lack a publicly accessible aerodynamics dataset that imitates real-world datasets. We compare the datasets used to highlight some of the differences between parametric and real-world datasets.    

We compare datasets using the Normalized Chamfer–Normal Distance (NCND), which combines a normalized point-wise Chamfer distance with nearest-neighbor normal cosine dissimilarity. Specifically, we calculate $\text{chamfer\_distance} / x + y * \text{cosine\_dissimilarity}$ between all pairs of geometries within a dataset. The constants $x$ and $y$ are chosen so the maximum possible normalized value is one and the largest contribution from each component is 0.5; the same constants are used across all plots for comparability. We compute the pairwise NCND distances within each dataset on 100,000-point clouds per geometry sampled via farthest point sampling to keep the computational cost reasonable. We calculate and report the distances in two settings: (i) Aligned geometries - meshes are first translated so that their centroid is at the origin and then shifted along the z-axis so that their minimum z is zero so that all models share a common ground plane; (ii) Scaled geometries - meshes are scaled to fit a unit cube, reducing proportional/placement effects and emphasizing feature-level differences. The scaled geometries tend to have major features (like wheels and mirrors) in similar locations. 

\Cref{fig:chamfer distance combined} shows the pairwise NCND distributions. The original real-world dataset geometries cover a broad span with significant irregularities. In comparison, although some parametric datasets, such as parametric I and DrivAerML, have a large span, they exhibit significantly lower irregularities compared to the real-world dataset. After scaling, average distances, range, and irregularities drop across all datasets. For all parametric datasets except the DrivAer dataset (mean reduction: 6\%; span reduction: 14\%), the mean drops by at least 40\% and the span by at least 48\%. Real-world also narrows (mean reduction: 20\%; span reduction: 28\%), but relative to the original geometry, its distribution stays the broadest and most irregular. This indicates that the real-world datasets have relatively more geometric changes that are not due to scaling or proportional changes. From the distribution calculated only for the samples within baseline groups (a group of geometries created from a single baseline geometry) for the real-world datasets, distances concentrate in a narrow region, indicating that most variation arises across baseline geometries rather than within a single baseline. This is shown to highlight that a majority of the changes in real-world geometries are targeted (as they do not significantly affect the NCND), specific changes that still considerably affect the $c_d$. The larger and significant proportional changes occur between the baseline geometries. Currently, all publicly available aerodynamic datasets are primarily parametric, and their diversity is driven mainly by large changes in a few geometric features. These datasets do not have the numerous diverse small geometric feature changes present in real-world datasets.

In \cref{fig:relative l2 pairwise chamfer distance combined} we compare the velocity fields by analyzing the distribution of the pairwise relative L2 distance computed within each dataset. Across all datasets, the distribution has a wide span. The distribution for all the parametric datasets yields a roughly normal distribution with varying means and standard deviations. In contrast, the real-world has a larger span with more irregularities and multiple peaks, underscoring that real-world flow fields vary more heterogeneously than the parametric datasets.

\section{Implementation details}

\subsection{PMRT}

\begin{figure}
  \centering
   \includegraphics[width=\linewidth]{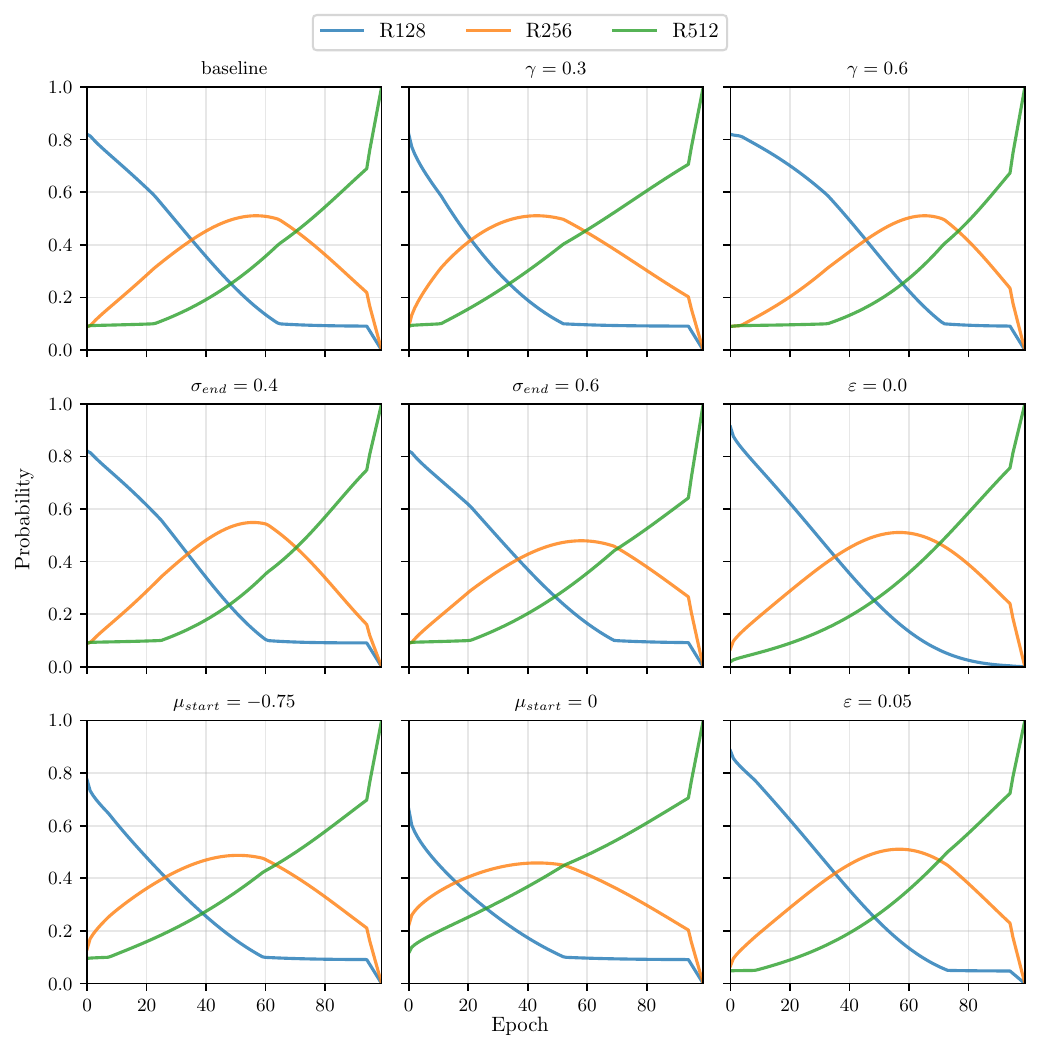}

   \caption{PMRT schedule some hyperparameters variations. Each subplot shows the sampling probabilities for three resolutions across the pre-training epochs with the subplot title showing the parameter that is changed relative to the baseline. The baseline configuration (top-left) uses $\gamma$=0.45, $\sigma_{\text{end}}$=0.5, $\epsilon$=0.1, and $\mu_{\text{start}}$=-1.5.}
    \label{fig:PMRT_schedule_grid}
\end{figure}

\PMRT consists of three phases: warm-up, pre-training, and fine-tuning. During warm-up, the probabilities are linearly interpolated from an equal distribution to the initial pre-training probabilities. During fine-tuning, the model trains exclusively on the highest resolution, \RHigh. The pre-training schedule is generated by discretizing a Gaussian distribution over the resolution indices, where over the pre-training epochs the mean moves from low toward higher resolutions while the standard deviation shrinks. Here, we provide the detailed implementation of the pre-training phase.

The pre-training schedule operates over a set of $R$ resolution levels, indexed by $r \in {0,\dots,R-1}$. In our case, $R=3$, corresponding to resolutions \RLow (0), \RMid (1), and \RHigh (2). The pre-training phase consists of $E$ epochs, indexed by $e \in {0,\dots,E-1}$. We first normalize the epoch index to the range $[0, 1]$:

\begin{align}
e_n &= \frac{e}{E-1}.
\label{eq:normalize_epoch}
\end{align}

The schedule is driven by mean ($\mu_e$) and standard deviation ($\sigma_e$) that varies over epochs, and the transition phase is controlled by $\gamma$:

\begin{align}
\mu_e &= \mu_{\text{start}} + \big(\mu_{\text{end}} - \mu_{\text{start}}\big)\, e_n^{\gamma}, \\
\sigma_e &= \sigma_{\text{start}} - \big(\sigma_{\text{start}} - \sigma_{\text{end}}\big)\, e_n^{\gamma}. \\
\mu_{\text{start}} &=-\tfrac{R}{2}, \quad
\mu_{\text{end}} = R-1, \quad
\sigma_{\text{start}} = \tfrac{R}{2}, \quad
\sigma_{\text{end}} = 0.5.
\label{eq:constants}
\end{align}

The hyperparameters $\mu_{\text{start}}$, $\mu_{\text{end}}$, $\sigma_{\text{start}}$,  $\sigma_{\text{end}}$ and $\gamma$ control the schedule. In our case, the mean moves from (-1.5 $\rightarrow$ 2) and the $\mu_{\text{start}}$ is set to $-R/2$, which gives a strong initial emphasis to the lowest resolution (\RLow). As the training progresses, the $\sigma$ shrinks and $\mu$ moves towards emphasizing the highest resolution. We use a $\gamma$ of 0.45 and $\sigma_\text{end}$ of 0.5 to ensure a balanced mixing of the resolutions. For example, with $E=200$ pre-training epochs, this leads to an overall sampling ratio of approximately 33\%, 35\%, and 32\% for resolutions \RLow, \RMid, and \RHigh. In our case, a higher $\gamma$ would give more emphasis to the lowest resolution and a lower $\gamma$ to the highest resolution. Similarly, a lower $\sigma_{\text{end}}$ would emphasize the \RHigh even more towards the end of the pre-training phase. \Cref{fig:PMRT_schedule_grid} shows how some hyperparameters affect the probabilities.

The raw probability for each resolution, $p^{\mathrm{raw}}_{e,r}$, is derived from the standard normal cumulative distribution function (CDF), $\Phi$ ($\Phi(x) = \tfrac{1}{2}\bigl(1 + \operatorname{erf}(x/\sqrt{2})\bigr)$), as formulated in \cref{eq:c0,eq:cdfedge,eq:praw}. In \cref{eq:ptilde,eq:prenorm}, the raw probabilities are clipped by a minimum value $\varepsilon$ (0.1) and normalized (sum to one) to prevent the probability of any resolution from getting too low. 

\begin{align}
c_{e,0} &= 0, \label{eq:c0}\\
c_{e,r+1} &= \Phi\!\left(\frac{(r+\tfrac{1}{2}) - \mu_e}{\sigma_e}\right) \label{eq:cdfedge}\\
p^{\mathrm{raw}}_{e,r} &= c_{e,r+1} - c_{e,r} \label{eq:praw}\\
\tilde{p}_{e,r} &= \max\!\big(p^{\mathrm{raw}}_{e,r},\, \varepsilon\big) \label{eq:ptilde}\\
p_{e,r} &= \frac{\tilde{p}_{e,r}}{\sum_{q=0}^{R-1} \tilde{p}_{e,q}} \label{eq:prenorm}
\end{align}

During the final $T$ epochs (5 epochs) of the pre-training phase (from epoch E-T to E-1), the sampling probabilities are linearly interpolated towards  the fine-tuning probabilities $p^{\mathrm{fine}}_r$, that select only the highest resolution:

\[
p_{E-T+t,\, r} = \Big(1 - \frac{t}{T-1}\Big)\, p_{E-T,\, r} \;+\; \frac{t}{T-1}\, p^{\mathrm{fine}}_r, \qquad t=0,\dots,T-1.
\]

Finally, during training, at each batch of a given epoch, a resolution is drawn from the distribution defined by the probabilities for that epoch:

\begin{align}
r_e \sim \operatorname{Categorical}\!\big(p_{e,0},\,p_{e,1},\,\dots,\,p_{e,R-1}\big)
\end{align}

\subsection{Performance}

One performance bottleneck we face is loading high-resolution data and applying augmentation. To mitigate these bottlenecks, we use MosaicML Streaming [\streamingcite] for data loading and Kornia [\korinacite] for data augmentation. 

All our models were trained using Float32. We have not experimented with mixed precision. All our models, excluding the \PMRT \RHigh $c_d$ + velocity model, do not saturate the GPU; for these models, we do not expect any significant performance gains using mixed precision except for the reduction in memory usage. It might be possible to train \RHigh \PMRT $c_d$ + velocity models at a lower compute cost with mixed precision.

\subsection{Region of interest (ROI)}
For the DrivAerML dataset, we extract the velocity fields in an ROI of size
\((L_x,L_y,L_z)=(\SI{9.28}{m},\,\SI{3.84}{m},\,\SI{2.66}{m})\). The ROI is anchored relative to the vehicle and CFD domain. The ROI starts \SI{1.0}{m} upstream of the vehicle, is laterally centered on the vehicle, and extends \SI{2.66}{m} vertically from the lower bound of the original CFD field (road plane). The \RHigh cell size (\(\approx\!\SI{0.0181}{m}\times\SI{0.03}{m}\times\SI{0.0208}{m}\)) is comparable to or finer than the original CFD cells except for some regions close to the vehicle's surface and in the wake of the car. While the CFD requires locally finer cells for a stable and accurate unsteady simulation, we find \RHigh sufficient for predicting and visualizing time-averaged velocity fields.

\section{Extended results}

\subsection{Ablations: architecture}

\begin{table*}
  \centering
  \begin{tabular*}{0.7\textwidth}{@{\extracolsep{\fill}}lcc@{}}
    \toprule
    \begin{tabular}[c]{@{}c@{}}Variant (relative to baseline)\end{tabular} &
    \begin{tabular}[c]{@{}c@{}}Group:\\ $c_d$ MAE $\downarrow$\end{tabular} &
    \begin{tabular}[c]{@{}c@{}}Group:\\ Relative L2\\ error $\downarrow$\end{tabular} \\
    \midrule
    Base: \RLow - $c_d$+velocity; 3 decoders                 & 2.4 & 2.7 \\
    Single velocity field decoder          & 2.5 & 3.3 \\
    No SDF to output convolution skip connection            & 2.7 & 2.8 \\
    Without multi-head self-attention and transformer & 2.7 & 2.7 \\
    \bottomrule
  \end{tabular*}
  \caption{Ablation of \RLow $c_d$ + velocity model. Rows indicate a changed feature, and we report the group $c_d$ MAE and group relative L2 error.}
  \label{tab: ablation arch}
\end{table*}

\Cref{tab: ablation arch} shows the ablations of the major architectural design choices. We compare all variants with the baseline \RLow $c_d$ + velocity model. When we replace the three velocity field decoders with a single decoder that predicts all three components, there is a negligible increase in the $c_d$ MAE but a significant increase in the relative L2 error. Removing the SDF to output convolution skip connection increases the $c_d$ MAE by 0.3 drag counts and the group relative L2 error by 0.1 drag counts. Removing the multi-head self-attention and transformer layers has no impact on the velocity field prediction but increases $c_d$ MAE by 0.3 drag counts. 

\subsection{Ablations: PMRT}

\begin{table*}
  \centering
  \begin{tabular*}{0.6\textwidth}{@{\extracolsep{\fill}}lc@{}}
    \toprule
    \begin{tabular}[c]{@{}c@{}}Model\end{tabular} &
    \begin{tabular}[c]{@{}c@{}}Group:\\ $c_d$ MAE $\downarrow$\end{tabular} \\
    \midrule
    Best \PMRT \RHigh                  & 2.5  \\
    \PMRT \RHigh: no probability floor                       & 2.6  \\
    \PMRT \RHigh: no warm-up                      & 2.7  \\
    \PMRT \RHigh: no fine-tuning on \RHigh                 & 3.0 \\
    \PMRT \RHigh: low resolution batch size multiplier of 2              & 2.8  \\
    \PMRT \RHigh: no low resolution batch size multiplier             & 3.0 \\
    Naive pre-train$\rightarrow$fine-tune (hard switch)                  & 3.3  \\
    Constant equal probability schedule & 3.6 \\
    \bottomrule
  \end{tabular*}
  \caption{Ablation of the \PMRT schedule for \RHigh $c_d$-only model. The best model uses a probability floor of 0.1, and a batch size multiplier of 4 for lower resolutions. Rows indicate the change compared to the best model. We report the group $c_d$ MAE.}
  \label{tab: ablation PRT}
\end{table*}

\Cref{tab: ablation PRT} reports ablations of the \PMRT schedule for the \RHigh $c_d$-only model, with all variants compared against the best \PMRT \RHigh model. Removing the probability floor or the warm-up has a minor increase in the $c_d$ MAE of 0.1 and 0.2 drag counts, respectively. Skipping the fine-tuning phase has a much larger increase of 0.5 drag counts on the $c_d$ MAE; as fine-tuning specializes the model to \RHigh. Reducing the low-resolution batch multiplier to $\times 2$ increased the group $c_d$ MAE by 0.3, and removing it entirely increases it by 0.5. The low-resolution batch size multiplier allows the model to see more low-resolution samples and improves the GPU utilization; the \RHigh batch size is memory-bound, so using the same batch size at lower resolutions underutilizes the GPU. In the naive pre-train$\rightarrow$ fine-tune, we first train on \RLow, then \RMid, and finally fine-tune it at \RHigh, which increases the group $c_d$ MAE by 0.8. The constant equal probability schedule samples all resolutions equally, which increases the group $c_d$ MAE by 1.1 drag counts. 

\subsection{Best model metrics in subgroups}

\Cref{tab:extended best model metrics} reports extended subgroup metrics for all the models. We include these for completeness; the interpretation of these metrics does not differ from what was interpreted using the aggregated metrics in the main paper.

\begin{table*}[t]
  \centering
  \begin{tabular*}{\textwidth}{@{\extracolsep{\fill}}llcccc@{}}
    \toprule
    & & \multicolumn{2}{c}{Baseline \RLow} & \multicolumn{2}{c}{Best \PMRT \RHigh} \\
    \cmidrule(r){3-4} \cmidrule(l){5-6}
    Dataset group & Metric & Without velocity & With velocity & Without velocity & With velocity \\
    \midrule
    DrivAerML & $c_d$: MAE $\downarrow$ & 3.7  & 2.5  & 2.8  & 2.4 \\
              & $c_d$: MaxAE $\downarrow$ & 17.9 & 15.8 & 12.7 & 8.9 \\
              & $c_d$: R$^2$ score $\uparrow$ & 0.911 & 0.958 & 0.951 & 0.969 \\
              & Velocity: Relative L2 error $\downarrow$ & -- & 2.9 & -- & 2.2 \\
    \midrule
    Parametric & $c_d$: MAE $\downarrow$ & 3.2  & 2.3  & 2.2  & 2.5 \\
               & $c_d$: MaxAE $\downarrow$ & 39.8 & 26.7 & 23.4 & 34.0 \\
               & $c_d$: R$^2$ score $\uparrow$ & 0.982 & 0.991 & 0.992 & 0.988 \\
               & Velocity: Relative L2 error $\downarrow$ & -- & 2.7 & -- & 2.6 \\
    \midrule
    Real-world & $c_d$: MAE $\downarrow$ & 3.2  & 2.5  & 2.6  & 2.8 \\
               & $c_d$: MaxAE $\downarrow$ & 29.5 & 10.1 & 17.7 & 26.2 \\
               & $c_d$: R$^2$ score $\uparrow$ & 0.900 & 0.961 & 0.950 & 0.915 \\
               & Velocity: Relative L2 error $\downarrow$ & -- & 2.4 & -- & 2.5 \\
    \midrule
    wall-clock time [\si{\hour}] & & 3 & 9 & 8 & 24 \\
    \bottomrule
  \end{tabular*}
    \caption{Extended subgroup test metrics for best models trained on all datasets. We report the MAE, MaxAE and $R^2$ score for $c_d$, and velocity relative L2 error when applicable. The final row reports wall-clock time on a single NVIDIA H100 GPU.}  \label{tab:extended best model metrics}
\end{table*}

\end{document}

%% file: Figures/src/overview_plot.tex
\begin{figure*}[htbp]
  \centering
   \includegraphics[width=0.99\linewidth]{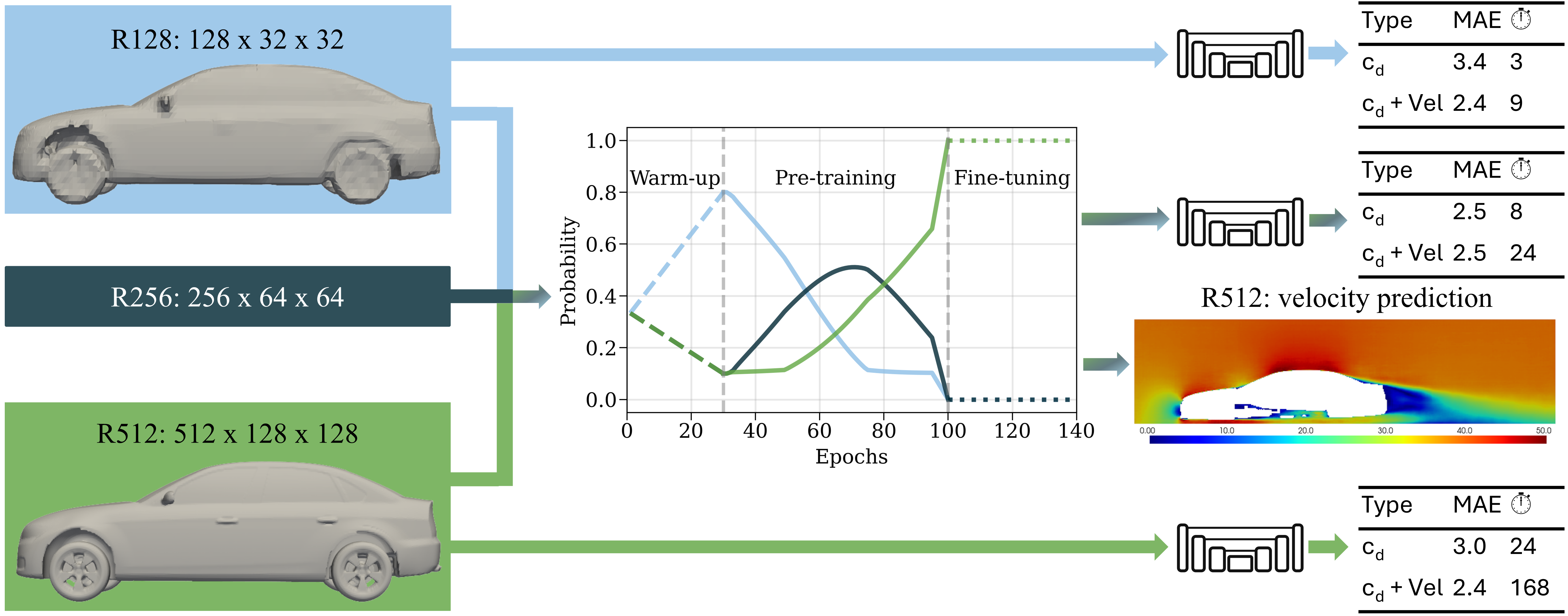}

   \caption{Overview of \PMRT and single-resolution baselines. Box colors, labeled with the resolution name (with example geometries reconstructed from SDFs for \RLow and \RHigh), represent the resolution's color used in the plot. Top/bottom: baselines trained only at low or high resolution. Middle: \PMRT pipeline; graph shows resolution probabilities over epochs for a hypothetical example, during training, batches are sampled based on these probabilities. Right: $c_d$ MAE and wall-clock time in hours for $c_d$-only and $c_d$+velocity variants with an example \PMRT \RHigh velocity field prediction.}
    \label{fig:prt_overview_plot}
\end{figure*}

%% file: Figures/src/drag_distribution.tex
\begin{figure}[t]
  \centering
   \includegraphics[width=0.99\linewidth]{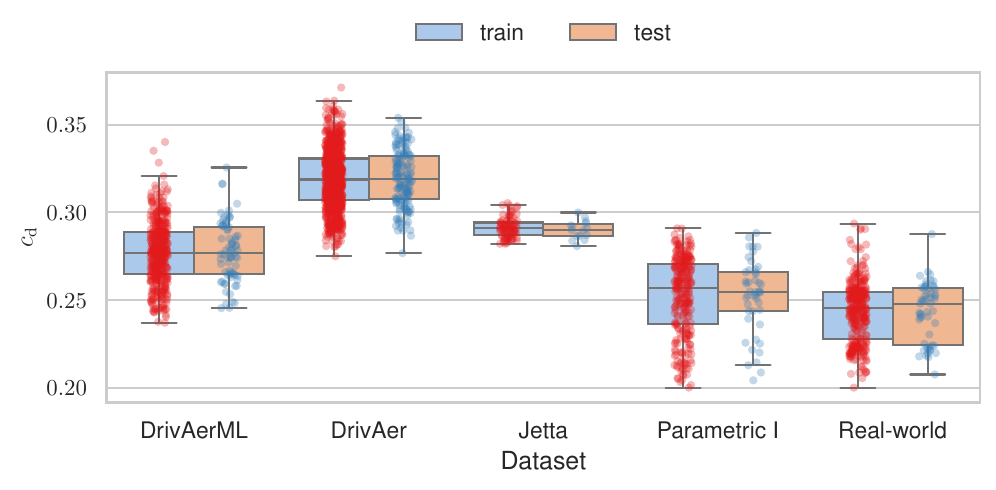}

   \caption{Distribution of $c_d$ across splits for all datasets. The $c_d$ values of the real-world dataset and Parametric I are subtracted by a constant such that the sample with the lowest $c_d$ has a value of 0.2 to protect sensitive $c_d$ data.}
   \label{fig:drag distribution}
\end{figure}

%% file: Figures/src/velocity_field_prediction_comparision.tex
\begin{figure*}[!htb]
  \centering
   \includegraphics[width=0.90\linewidth]{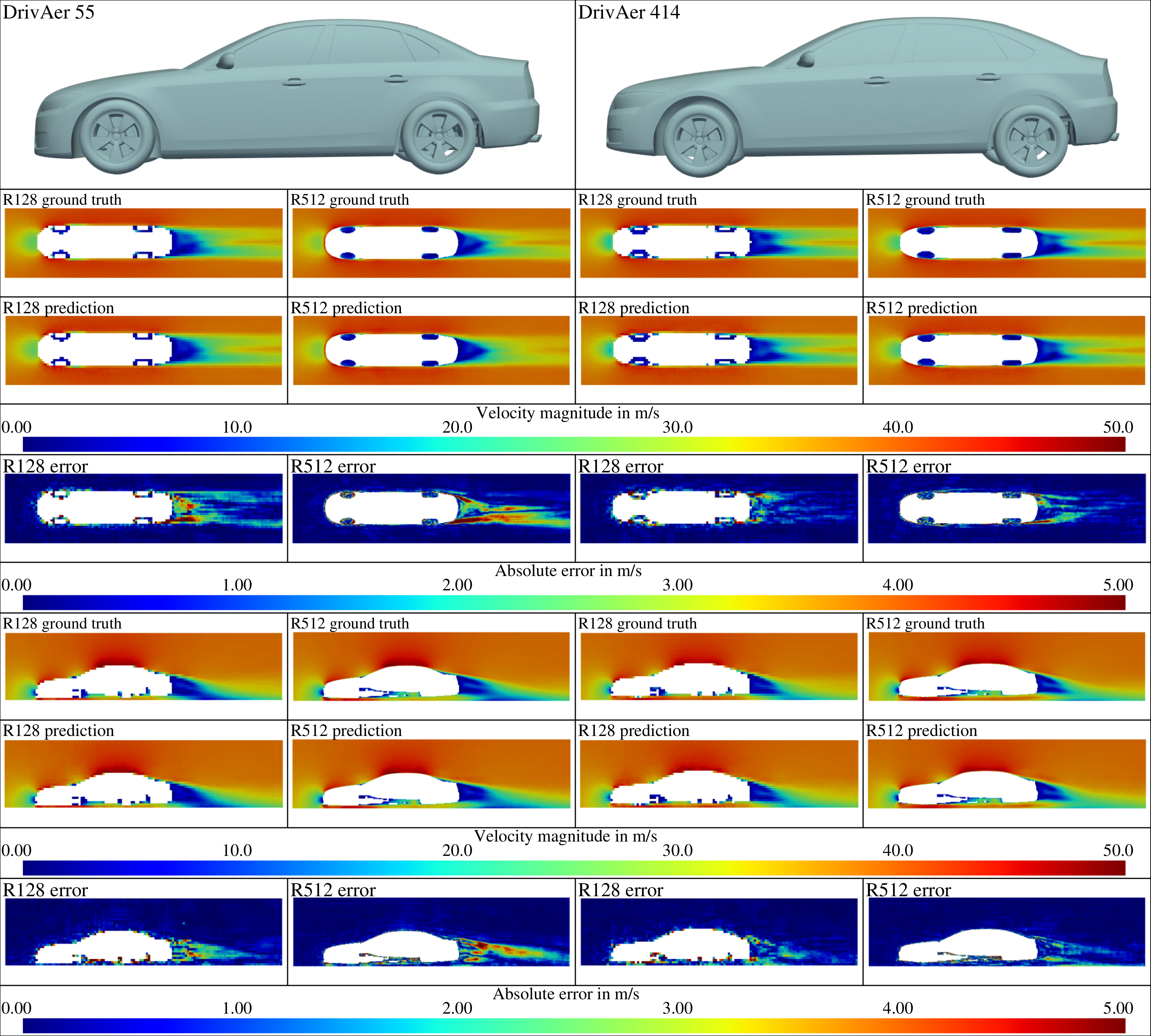}

   \caption{Comparison of true and predicted velocity field slices between \RLow and \PMRT \RHigh models for two DrivAerML samples.}
   \label{fig:velocity field comparision}
\end{figure*}

%% file: Figures/src/drag_correlation_plot.tex
\begin{figure*}[h]
  \centering
   \includegraphics[width=0.90\linewidth]{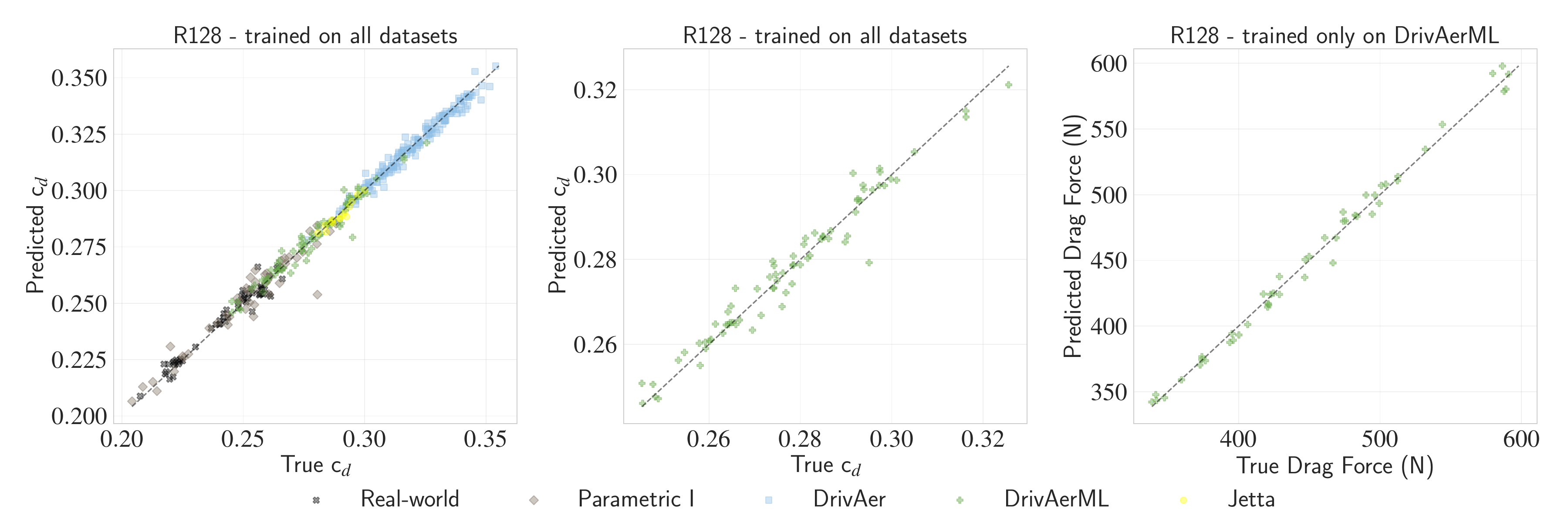}

   \caption{Correlation plots for \RLow\ models predicting \(c_d\) and velocity: (1) \(c_d\) on all datasets; (2) \(c_d\) on DrivAerML (model trained on all datasets); (3) drag force on DrivAerML with a model trained only on DrivAerML using benchmark split.}
    \label{fig:correlation plot}
\end{figure*}

%% file: Figures/src/dataset_comparision.tex
\begin{figure*}[h]
  \centering
   \includegraphics[width=\linewidth]{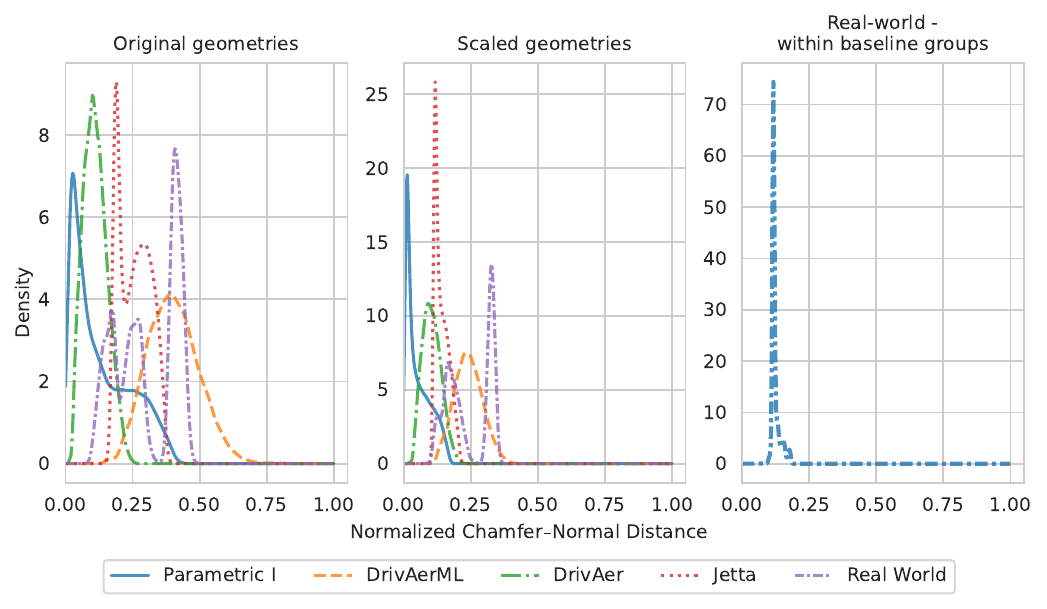}

   \caption{Normalized Chamfer–Normal Distance (NCND) distribution of pairwise samples within a dataset. A larger NCND distance indicates a larger difference between the geometries. The plots compare the NCND distribution in order from left to right: (1) aligned geometries: geometries translated to a common reference; (2) scaled geometries: geometries that are scaled to a unit box to reduce the effect of proportional changes; (3) pairwise distances of real-world dataset calculated only with baseline groups.}
    \label{fig:chamfer distance combined}
\end{figure*}

\begin{figure*}[htbp]
  \centering
   \includegraphics[width=\linewidth]{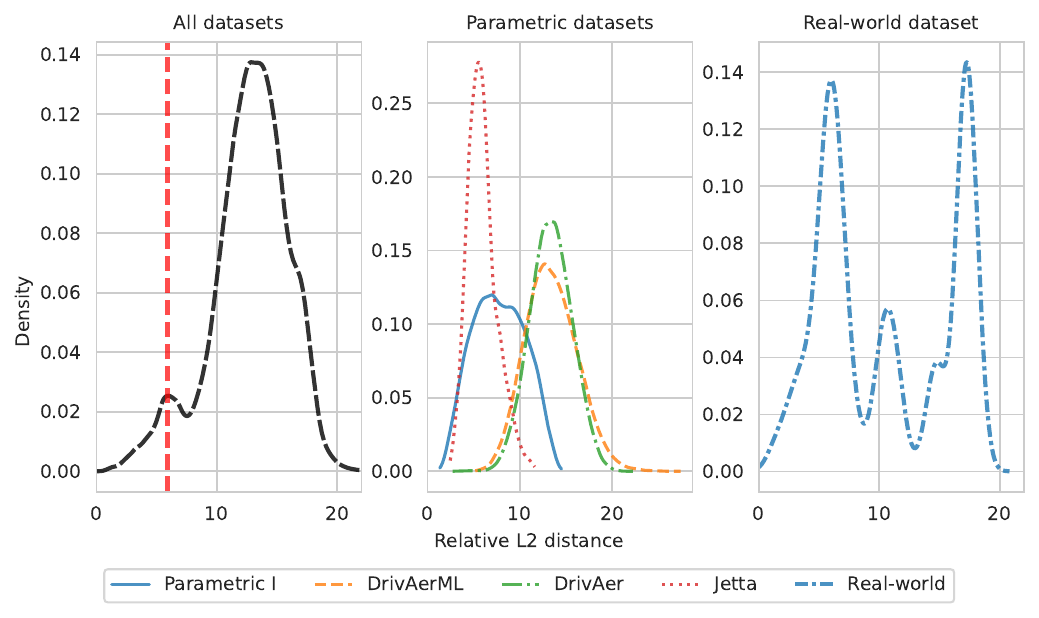}

   \caption{Distribution of pairwise relative L2 distance between time-averaged velocity fields. The plots, left to right, show: (1) all datasets combined: with per-dataset distributions computed from all within-dataset pairs; the vertical red line marks the error that would be obtained by a nearest-neighbor baseline that, for each test sample, predicts the velocity field of its closest training sample; (2) only parametric datasets (3) only the real-world dataset. Larger values indicate larger flow-field differences.}
    \label{fig:relative l2 pairwise chamfer distance combined}
\end{figure*}